\documentclass[12pt]{article} 
\usepackage{amsmath}
\usepackage{graphicx}

\renewcommand{\baselinestretch}{1.3}

\begin{document}

\begin{center}

{\large \bf Optimal statistical decision for Gaussian graphical model selection}

Valery A. Kalyagin (a),  Alexander P. Koldanov (a), \\
Petr A. Koldanov (a), Panos M. Pardalos (a, b)
\footnote{(a) - Laboratory of Algorithms and Technologies for Network Analysis, National Research University
Higher School of Economics, Bolshaya Pecherskaya 25/12, Nizhny
Novgorod, 603155 Russia,  {\it vkalyagin@hse.ru} and (b) - University of Florida, ISE Department, Gainesville, FL 32611, USA.}

\end{center}



\noindent
{\bf Abstract.} Gaussian graphical model is a graphical representation of the dependence structure for a Gaussian random vector. It is recognized as a powerful tool in different applied fields such as bioinformatics, error-control codes, speech language, information retrieval and others. Gaussian graphical model selection is a statistical problem to identify the Gaussian graphical model from a sample of a given size. Different approaches for Gaussian graphical model selection are suggested in the literature. One of them is  based on considering  the family of individual conditional independence tests. The application of this approach leads to the construction of a variety of multiple testing statistical procedures for Gaussian graphical model selection. An important characteristic of these procedures is its error rate for a given sample size. In existing literature great attention is paid to the control of error rates for incorrect edge inclusion (Type I error). However, in graphical model selection it is also  important to take into account error rates for incorrect edge exclusion (Type II error).  To deal with this issue we  consider the graphical model selection problem in the framework of the multiple decision theory. The quality of statistical procedures is measured by a risk function with additive losses. Additive losses allow both types of  errors to be taken into account.  We construct  the tests of a Neyman structure for individual hypotheses and combine them to obtain a multiple decision statistical procedure. We show that the obtained procedure is optimal in the sense that it minimizes the linear combination of expected numbers of Type I and Type II errors in the class of unbiased multiple decision procedures. 


\renewcommand{\baselinestretch}{1.6}

\section{Introduction}
Gaussian graphical model is a graphical representation of the dependence structure for a  Gaussian random vector. It is recognized as a powerful tool in different applied fields such as bioinformatics, error-control codes, speech language, information retrieval and others \cite{Jordan_2004}. There are two aspects related to the study of the graphical model: the computational and  algorithmic aspect, and the statistical aspect. The computational aspect has become increasingly popular  due to the growing interest in large scale networks \cite{Jordan_2008}. A central question in the statistical aspect  is how to recover the structure of an undirected Gaussian graph from observations. This problem is called the Gaussian graphical model selection problem (GGMS problem). A comprehensive survey on different approaches to this problem is given in \cite{Drton_2007}. Most of the constructed statistical procedures for GGMS are based on asymptotically optimal estimations of correlations  \cite{Anderson_2003}, \cite{Drton_2004}, \cite{Zhao_Ren_2015}. However, as far as we know, there are no results related to the optimality of statistical procedures 
for GGMS for a fixed sample size. This is the subject of this paper. 

The quality of  statistical procedures for GGMS can be measured by two types of errors: false edge inclusion (Type I error) and false edge exclusion (Type II error). Traditional measures of quality used in GGMS are: FWER (Family Wise Error Rate),  FDR (False Discovery Rate), FDP (False Discovery Proportion) and others \cite{Drton_2007}.  Most of them are connected with the Type I error (false edge inclusion). It is clear, that the quality of GGMS procedures has to be related to the difference between two graphs: the true graph and selected graph. Therefore in GGMS it is important to take into account both types of errors (Type I and Type II errors). The quality of a statistical procedure can be measured in this case by the linear combination of expected values of the numbers of Type I and Type II errors.  We refer to  a selection statistical procedure which minimizes this value for a fixed sample size as {\em optimal}. The main goal of this paper is to find an optimal  statistical procedure for GGMS. To achieve this goal we  consider the graphical model selection problem as part of the multiple decision theory \cite{Lehmann_1957}. The quality of statistical procedures is measured by the risk function.  To take into account the numbers of Type I and Type II errors we consider the additive loss function.  We prove  that in this case the risk function is a linear combination of expected values of the numbers of Type I and Type II errors. Subsequently, we construct  tests of a Neyman structure  for individual hypotheses and combine them to obtain a multiple decision statistical procedure. We show that the obtained procedure minimizes the risk function in the class of unbiased multiple decision procedures. 

The paper is organized as follows. In  Section \ref{Problem statement} we give  basic definitions and notations. 
In  Section \ref{Multiple decision approach} we describe the multiple decision framework for the Gaussian graphical model selection problem and prove the representation of the risk function. 
In Section \ref{UMPU test for individual hypotheses} we construct the tests of a Neyman structure  for individual hypotheses.  
In Section \ref{Optimal multiple decision procedures} we combine individual tests in the multiple decision procedure and prove its optimality. 
In Section \ref{Concluding remarks} we discuss the proposed approach for different model selection problems.

\section{Problem statement}\label{Problem statement}
Let $X=(X_1,X_2,\ldots,X_p)$ be a random vector with the multivariate Gaussian distribution from $N(\mu, \Sigma)$, where $\mu=(\mu_1,\mu_2,\ldots, \mu_p)$ is the vector of means and 
$\Sigma=(\sigma_{i,j})$ is the covariance matrix, $\sigma_{i,j}=\mbox{cov}(X_i,X_j)$, $i,j=1,2,\ldots,p$. Let $x(t)$, $t=1,2,\ldots,n$  be a sample of size $n$ from the distribution of $X$. 
We assume in this paper that $n>p$, and that the matrix $\Sigma$ is non degenerate. The case $n<p$ has a practical interest too \cite{Liang_2015}, but it is not considered in this paper.   
The undirected Gaussian graphical model is an undirected graph with $p$ nodes. The nodes of the graph are associated with the random variables $X_1,X_2,\ldots,X_p$, edge $(i,j)$ is included in the graph if the random 
variables $X_i,X_j$ are conditionally dependent \cite{Lauritzen_1996}, \cite{Anderson_2003}.  Gaussian graphical model selection problem consists of the identification of a graphical model from observations. 
 
The partial correlation $\rho_{i,j \bullet N(i,j)}$ of $X_i$, $X_j$ given $X_k$, $k \in N(i,j) = \{1,2,\ldots,p\}\setminus \{i,j\}$ is defined as the correlation of $X_i$, $X_j$ 
in the conditional distribution of $X_i$, $X_j$ given $X_k$, $k \in N(i,j)$. It is known\cite{Anderson_2003} that the conditional distribution of $X_i$, $X_j$ given 
$X_k$, $k \in N(i,j)$ is Gaussian with the correlation $\rho_{i,j \bullet N(i,j)}$. It implies that the conditional independence of $X_i$, $X_j$ given $X_k$, 
$k \in N(i,j) = \{1,2,\ldots,p\}\setminus \{i,j\}$ is equivalent to the equation $\rho_{i,j \bullet N(i,j)}=0$. Therefore, the Gaussian graphical model selection is equivalent 
to simultaneous inference on hypotheses of pairwise conditional independence $\rho_{i,j \bullet N(i,j)}=0$, $i \neq j$, $i,j=1,2,\ldots,p$. 

The inverse matrix for $\Sigma$,  $\Sigma^{-1}=(\sigma^{i,j})$ is known as the concentration or precision matrix for the distribution of $X$.   
For simplicity we use the notation $\rho^{i,j}=\rho_{i,j \bullet N(i,j)}$. The problem of pairwise conditional independence testing has the form:
\begin{equation}\label{main_problem}
h_{i,j}:\rho^{i,j}=0 \mbox{ vs } k_{i,j}:\rho^{i,j}\neq 0, \quad i \neq j, i,j=1,2,\ldots,p
\end{equation}
According to \cite{Lauritzen_1996} the partial correlation can be written as  
$$
\rho^{i,j}=-\frac{\sigma^{i,j}}{\sqrt{\sigma^{i,i}\sigma^{j,j}}}
$$
Note that the problem of pairwise conditional independence testing (\ref{main_problem}) is equivalent to 
$$
h_{i, j}:\sigma^{i,j}=0, \  \mbox{ vs } \  k_{i, j}:\sigma^{i,j}\neq 0, i \neq j, i,j=1,2,\ldots,p
$$
The Gaussian graphical model selection problem can be formulated now as multiple testing problem for the set of hypotheses (\ref{main_problem}).

\section{Multiple decision approach}\label{Multiple decision approach}

In this Section we consider the GGMS problem in the framework of  decision theory \cite{Wald_1950}. According to this approach we specify the decision statistical procedures  and risk function. 
Let $X=(X_1,X_2,\ldots,X_p)$ be a random vector with multivariate Gaussian distribution from $N(\mu, \Sigma)$. In the GGMS study observations are modeled as a sequence of random vectors $X(t)$, $t =1, 2,\ldots,n$
where $n$ is the sample size and vectors $X(t)$ are independent and identically distributed as $X$. Let $x=(x_{i}(t))$ be  observations of the random variables $X_i(t)$, $t =1, 2, \ldots,n$, $i = 1,2, \ldots,p$. 
Consider the set $\cal{G}$ of all $p \times p$ symmetric matrices  $G=(g_{i,j})$ with $g_{i,j} \in \{0,1\}$, $i,j=1,2,\ldots,p$, $g_{i,i}=0$, $i=1,2,\ldots,p$. Matrices $G \in \cal{G}$ represent adjacency 
matrices of all simple undirected graphs with $p$ vertices. The total number of matrices in $\cal{G}$ is equal to $L=2^M$ with $M=p(p-1)/2$. 
The {\it GGMS problem} can be formulated as a multiple decision problem of the choice between $L$ hypotheses:
\begin{equation}\label{N hypotheses}
H_G: \rho^{i,j}=0, { if } g_{i,j}=0, \ \ \rho^{i,j} \neq 0 \mbox{ if } g_{i,j}=1; \ \ i \neq j, \ \ i,j = 1,2,\ldots, p
\end{equation}
The multiple decision statistical procedure $\delta(x)$ is a map from the sample space $R^{p \times n}$ to the decision space $D=\{d_G, g \in \cal{G} \}$,  where the decision  $d_G$ is the
acceptance of hypothesis $H_G$, $G \in \cal{G}$. 
Let $\varphi_{i,j}(x)$ be  tests for the individual hypothesis (\ref{main_problem}). More precisely, 
$\varphi_{i,j}(x)=1$ means that hypothesis $h_{i,j}$ is rejected (edge $(i,j)$ is included in the graphical model), and $\varphi_{i,j}(x)=0$ means that hypothesis $h_{i,j}$ is accepted (edge $(i,j)$ is not included in the graphical model). 
Let $\Phi(x)$ be the matrix
\begin{equation}\label{test_for_N_hypotheses_overall_form}
\Phi(x)=\left(\begin{array}{cccc}
0 &\varphi_{1, 2}(x) &\ldots &\varphi_{1, p}(x)\\
\varphi_{2, 1}(x) & 0 &\ldots &\varphi_{2, p}(x)\\
\ldots&\ldots&\ldots&\ldots\\
\varphi_{p, 1}(x) &\varphi_{p, 2}(x) &\ldots & 0\\
\end{array}\right).
\end{equation}
Any multiple decision statistical procedure $\delta(x)$  based on the simultaneous inference of individual edge tests (\ref{main_problem}) can be written as
\begin{equation}\label{mdp_overall_form}
\delta(x)=d_G, \  \mbox{iff} \  \Phi(x)=G
\end{equation}

According to \cite{Wald_1950} the quality of the statistical procedure 
is defined by the risk function. Let $\Omega$ be the set of parameters $\Omega=\{\theta: \theta=(\mu, \Sigma), \mu \in R^p$, $\Sigma \mbox{ is a symmetric positive definite  matrix} \}$. 
By $\Omega_S$ we denote the parametric region corresponding to hypothesis $H_S$. 
Let $S=(s_{i,j})$, $Q=(q_{i,j})$, $S$, $Q$ $\in \cal{G}$. By $w(S,Q)$ we denote the loss from decision $d_Q$ when hypothesis $H_S$ is true, i.e.
$$
w(H_S;d_Q)=w(S,Q), \ \ S,Q \in \cal{G}
$$ 
Assume that $w(S,S)=0, S \in \cal{G}$. The risk function is defined by
$$
Risk(S, \theta;\delta)=\sum_{Q \in \cal{G}} w(S,Q)P_{\theta}(\delta(x)=d_Q), 
$$
where $P_{\theta}(\delta(x)=d_Q)$ is the probability that decision $d_Q$ is taken. 

As mentioned before, for the GGMS problem it is important to control Type I and Type II errors. 
Let $a_{i,j}$ be the loss from the false inclusion of edge $(i,j)$ in the graphical model, and let $b_{i,j}$, be the
loss from the false non inclusion of the edge $(i,j)$ in the graphical model, $i,j=1,2,\ldots,p; \ i\neq j$.  

Define the individual loss as
$$w_{i,j}(S,Q)=\left\{\begin{array}{cc}
a_{i,j}, & \mbox{if } \ s_{i,j}=0,q_{i,j}=1, \\
b_{i,j}, & \mbox{if } \ s_{i,j}=1,q_{i,j}=0, \\
0, & \mbox{ otherwise }
\end{array}\right.$$  

To take into account both types of errors we suggest the  total loss  $w(S,Q)$ is defined as:
\begin{equation}\label{additive_loss_function}
w(S,Q)=\sum_{i=1}^p\sum_{j=1}^p w_{i,j} (S,Q)
\end{equation}
It means that the total loss from the misclassification of $H_S$ is equal to the sum of losses from the misclassification of individual edges:
$$
w(S,Q)=\sum_{\{i,j:s_{i,j}=0;q_{i,j}=1\}}a_{i,j}+\sum_{\{i,j:s_{i,j}=1;q_{i,j}=0\}}b_{i,j} 
$$
The main result of this Section is the following theorem
 
\newtheorem{teo}{Theorem}
\begin{teo}
Let the loss function $w$ be defined by (\ref{additive_loss_function}), and $a_{i,j}=a$, $b_{i,j}=b$, $i \neq j$, $i,j=1,2,\ldots,p$. Then 
$$
Risk(S, \theta;\delta)=aE_{\theta}[Y_I(S, \delta)]+bE_{\theta}[Y_{II}(S, \delta)]
$$
where $Y_I(S, \delta)$, $Y_{II}(S, \delta)$ are the numbers of Type I and Type II errors for model selection by the statistical procedure $\delta$ when the true decision is $d_S$. 
\end{teo}

\noindent
{\bf Proof.} One has
$$
Risk(S, \theta;\delta)=\sum_{Q \in \cal{G}} w(S,Q)P_{\theta}(\delta(x)=d_Q)=
$$
$$
\sum_{Q \in \cal{G}}[\sum_{\{i,j:s_{i,j}=0;q_{i,j}=1\}}a_{i,j}+\sum_{\{i,j:s_{i,j}=1;q_{i,j}=0\}}b_{i,j}]P_{\theta}(\delta(x)=d_Q)=
$$
$$
= \sum_{Q \in \cal{G}} [aN_I(Q)+bN_{II}(Q)]P_{\theta}(\delta(x)=d_Q)=aE_{\theta}[Y_I(S, \delta]+bE_{\theta}[(Y_{II}(S, \delta)]
$$
where $N_I(Q)$ is the number of Type I errors when the procedure $\delta(x)$ takes decision $d_Q$  and $\theta \in \Omega_S$, and $N_{II}(Q)$ is the number of Type II errors 
when the  procedure $\delta(x)$ takes decision $d_Q$ and $\theta \in \Omega_S$.

\section{Uniformly most powerful unbiased tests for individual hypotheses}\label{UMPU test for individual hypotheses}
In this Section we briefly present the uniformly most powerful unbiased tests for individual hypotheses  (\ref{main_problem}). More details are given in our paper \cite{Koldanov_2017}. 
Consider the statistics 
$$
S_{k,l}=\frac{1}{n} \Sigma_{t=1}^n(X_{k}(t)-\overline{X_{k}})(X_{l}(t)-\overline{X_{l}}),
$$
The joint distribution  of statistics $S_{k,l}$, $k,l = 1,2,\ldots,N$, $n>p$ is given by the Wishart density function \cite{Anderson_2003}:
$$
f(\{s_{k,l}\})=\displaystyle \frac{ [\det (\sigma^{k,l})]^{n/2}
\times [\det(s_{k,l})]^{(n-p-2)/2}\times \exp[-(1/2)\sum_k \sum_l
s_{k,l} \sigma^{k,l}]} {2^{(pn/2)}\times \pi^{p(p-1)/4} \times
\Gamma(n/2)\Gamma((n-1)/2)\cdots\Gamma((n-p+1)/2)}
$$
if the  matrix $S=(s_{k,l})$ is positive definite, and $f(\{s_{k,l}\})=0$ otherwise. The  Wishart density function can be written as:
$$
f(\{s_{k,l}\})=\displaystyle C(\{\sigma^{k,l}\}) 
\exp[-\sigma^{i,j}s_{i,j} - \frac{1}{2} \sum_{(k,l)\neq
(i,j);(k,l)\neq(j,i)} s_{k,l} \sigma^{k,l}]  m(\{s_{k,l}\})
$$
where
$$
C(\{\sigma^{k,l}\})=c_1^{-1}[\det (\sigma^{k,l})]^{n/2}
$$
$$
c_1=2^{(pn/2)}\times \pi^{p(p-1)/4}\times
\Gamma(n/2)\Gamma((n-1)/2)\cdots\Gamma((n-p+1)/2)
$$
$$ m(\{s_{k,l}\})=[\det(s_{k,l})]^{(n-p-2)/2}
$$

According to \cite{Lehmann_2005} (Ch. 4)  the uniformly most powerful unbiased (UMPU) test for hypothesis $h_{i,j}$ has the form:
\begin{equation}\label{Nstructure}
\varphi_{i, j}(\{s_{k, l}\})=\left\{\begin{array}{rl}
 \ 0, &\mbox{}\: if \:  c_{i,j}'(\{s_{k,l}\})<s_{i,j}<c_{i,j}'' (\{s_{k,l}\}),\  (k,l)\neq (i,j)\\
 \ 1, &\mbox{}\: if \: s_{i,j}\leq c_{i,j}'(\{s_{k,l}\})\mbox{ or } s_{i,j}\geq c_{i,j}''(\{s_{k,l}\}),\  (k,l)\neq (i,j)
 \end{array}\right.
\end{equation}
where  the critical values $c'_{i,j}, c''_{i,j}$  are defined from the equations 
\begin{equation}\label{threshold1}
\displaystyle \frac{\int_{I \cap [c_{i,j}';c_{i,j}'']}
 [\det(s_{k,l})]^{(n-p-2)/2}  ds_{i,j}}
{\int_{I}  [\det(s_{k,l})]^{(n-p-2)/2}
ds_{i,j}} =1-\alpha_{i,j}
\end{equation}
\begin{equation}\label{threshold2}
\begin{array}{l}
\displaystyle \int_{I \cap (-\infty;c_{i,j}']}
s_{i,j}[\det(s_{k,l})]^{(n-p-2)/2}
ds_{i,j}+\\
+\displaystyle \int_{I \cap [c_{i,j}'';+\infty)}
s_{i,j} [\det(s_{k,l})]^{(n-p-2)/2}
ds_{i,j}=\\
 =\alpha_{i,j} \int_I s_{i,j}[\det(s_{k,l})]^{(n-p-2)/2} ds_{i,j}
\end{array}
\end{equation}
where $I$ is the interval of values of $s_{i,j}$ such that the matrix $S=(s_{k,l})$ is positive definite, and  $\alpha_{i,j}$ is the  significance level of the test.

It is shown in \cite{Koldanov_2017} that the constructed UMPU test is equivalent to the following partial correlation test
\begin{equation}\label{Neyman_structure_final}
\varphi_{i,j}^{umpu}=\left\{\ 
\begin{array}{ll} 
0, & \displaystyle 2q_{i,j}-1 < r^{i,j} < 1-2q_{i,j} \\
1, & \mbox{otherwise}, 
\end{array}\right.
\end{equation}
where $r^{i,j}$ is the sample partial correlation, and $q_{i,j}$ is the $(\alpha_{i,j}/2)$-quantile of the beta distribution $Be(\frac{n-p}{2},\frac{n-p}{2})$. 

Finally, we need to specify the optimality and unbiasedeness of the test  (\ref{Neyman_structure_final}). Let $\omega_{i,j}$ be the set of parameters defined by
$$
\omega_{i,j}=\bigcup _{S: s_{i,j}=0}\Omega_S
$$
Denote by $\omega_{i,j}^{-1}=\Omega \setminus w_{i,j}$. Let $\varphi_{i,j}$ be a test for individual hypothesis $h_{i,j}$ with significance level $\alpha_{i,j}$. 
The test $\varphi_{i,j}$ is reffered to as unbiased \cite{Lehmann_2005} if  
\begin{equation}\label{unbiased_individual}
E_{\theta}(\varphi_{i,j}) \leq \alpha_{i,j}, \ \forall \theta \in \omega_{i,j}; \ \ 
E_{\theta}(\varphi_{i,j}) \geq \alpha_{i,j}, \ \forall \theta \in \omega_{i,j}^{-1}
\end{equation}
The test $\varphi_{i,j}^{umpu}$ defined by (\ref{Neyman_structure_final}) is unbiased and the following inequality holds:
\begin{equation}\label{umpu_individual}
E_{\theta}(\varphi_{i,j}^{umpu}) \geq E_{\theta}(\varphi_{i,j}), \ \ \forall \theta \in \omega_{i,j}^{-1} 
\end{equation}
for any unbiased test $\varphi_{i,j}$.

\section{Optimal multiple decision procedures}\label{Optimal multiple decision procedures}

According to Wald \cite{Wald_1950} the procedure $\delta^*$ is referred to as optimal in the class of statistical procedures $\cal{C}$ if
\begin{equation}\label{optimality}
Risk(S, \theta; \delta^*) \leq Risk(S, \theta; \delta), 
\end{equation} 
for any $S \in \cal{G}$, $\theta \in {\Omega}_S$, $\delta \in \cal{C}$. 

In this paper we consider the class of $w$-unbiased statistical procedures. Statistical procedure $\delta(x)$ is reffered to as  $w$-unbiased if one has
\begin{equation}\label{unbiasedeness}
Risk(S, \theta;\delta)=E_{\theta} w(S; \delta) \leq E_{\theta} w(S'; \delta)=Risk(S', \theta;\delta), 
\end{equation}
for any $S, S' \in \cal{G}$, $\theta \in {\Omega}_S$. The following theorem describes the optimal  procedure in the class of $w$-unbiased multiple decision procedures for Gaussian graphical model selection.  

\begin{teo}
Let the loss function $w$ be defined by (\ref{additive_loss_function}). Let the procedure $\delta^{ou}$  be defined by (\ref{test_for_N_hypotheses_overall_form})- (\ref{mdp_overall_form}), where $\varphi_{i,j}=\varphi_{i,j}^{umpu}$ is defined by (\ref{Neyman_structure_final}) and 
\begin{equation}\label{abalpha}
\alpha_{i,j}=\frac{b_{i,j}}{a_{i,j}+b_{i,j}}.
\end{equation}
Then procedure $\delta^{ou}$ is an optimal multiple decision statistical procedure in the class of $w$-unbiased procedures for Gaussian graphical model selection. 
\end{teo}

\noindent
{\bf Proof.} We use the general approach by Lehmann  \cite{Lehmann_1957} and give a direct proof. 
Let $\delta$ be a statistical procedure defined by (\ref{test_for_N_hypotheses_overall_form})-(\ref{mdp_overall_form}). If the loss function $w$ satisfies (\ref{additive_loss_function})  then the risk of statistical procedure $\delta$ is the sum of the risks of individual tests $\varphi_{i,j}$. Indeed, one has 
$$
Risk(S, \theta;\delta)=\sum_{Q \in \cal{G}} w(S,Q)P_{\theta}(\delta(x)=d_Q)=
$$
$$
= \sum_{Q \in \cal{G}}[\sum_{\{i,j:s_{i,j}=0;q_{i,j}=1\}}a_{i,j}+\sum_{\{i,j:s_{i,j}=1;q_{i,j}=0\}}b_{i,j}]P_{\theta}(\delta(x)=d_Q)=
$$
$$
= \sum^p_{i, j=1, s_{i,j}=0} a_{i,j} \sum_{Q, q_{i,j}=1} P_{\theta}(\delta(x)=d_Q)+
\sum^p_{i, j=1, s_{i,j}=1} b_{i,j} \sum_{Q, q_{i,j}=0} P_{\theta}(\delta(x)=d_Q)=
$$
$$
=\sum^p_{i, j=1, s_{i,j}=0}a_{i,j}P_{\theta}(\varphi_{i,j}(x)=1)+\sum^p_{i, j=1, s_{i,j}=1} b_{i,j}P_{\theta}(\varphi_{i,j}(x)=0)=
$$
$$
=\sum_{i=1}^p\sum_{j=1}^p Risk(s_{i,j},\theta; \varphi_{i,j})
$$
where
$$
Risk(s_{i,j}, \theta; \varphi_{i,j})=\left\{\begin{array}{lll}
a_{i,j}P_{\theta}(\varphi_{i,j}=1), & \mbox{if} & \theta \in \omega_{i,j} \\ 
b_{i,j}P_{\theta}(\varphi_{i,j}=0), & \mbox{if} & \theta \in \omega_{i,j}^{-1} \\  
\end{array}\right.
$$

Now we prove that $\delta^{ou}$ is a $w$-unbiased multiple decision procedure. One has from relation (\ref{abalpha})  and unbiasedness (\ref{unbiased_individual}) of $\varphi^{umpu}$ 
$$
a_{i,j}P_{\theta}(\varphi^{umpu}{i,j}=1) \leq b_{i,j}P_{\theta}(\varphi^{umpu}_{i,j}=0), \ \mbox{if} \ \theta \in \omega_{i,j},  
$$
and
$$
b_{i,j}P_{\theta}(\varphi^{umpu}{i,j}=0) \leq a_{i,j}P_{\theta}(\varphi^{umpu}_{i,j}=1), \ \mbox{if} \ \theta \in \omega^{-1}_{i,j} 
$$
It implies that
$$
Risk(s_{i,j}, \theta; \varphi_{i,j}) \leq Risk(s'_{i,j}, \theta; \varphi_{i,j}), \ \ \forall s_{i,j}, s'_{i,j}=0,1
$$
Therefore,
$$
Risk(S, \theta;\delta^{ou})= \sum_{i=1}^p\sum_{j=1}^p Risk(s_{i,j},\theta; \varphi^{umpu}_{i,j}) \leq 
$$
$$
\leq \sum_{i=1}^p\sum_{j=1}^p Risk(s'_{i,j},\theta; \varphi^{umpu}_{i,j})=Risk(S', \theta;\delta^{ou}), 
$$
for any $S, S' \in \cal{G}$, $\theta \in {\Omega}_S$. 

Finally, we prove that $\delta^{ou}$ is optimal in the class of $w$-unbiased statistical procedures.
Let $\delta$ be an $w$-unbiased statistical procedure defined by (\ref{test_for_N_hypotheses_overall_form})-(\ref{mdp_overall_form}). 
One has: 
$$
Risk(S, \theta;\delta) \leq Risk(S', \theta;\delta), \ \ \forall S,S' \in \cal{G}
$$
Take $S'$ such that $S'$ and $S$ differ only in two positions $(i,j)$ and $(j,i)$. In this case one has
$$
Risk(S, \theta;\delta)= 2Risk(s_{i,j},\theta; \varphi_{i,j}) +\sum_{(k,l) \neq (i,j)} Risk(s_{k,l},\theta; \varphi_{k.l})
$$
and:
$$
Risk(S', \theta;\delta)= 2Risk(s'_{i,j},\theta; \varphi_{i,j}) +\sum_{(k,l) \neq (i,j)} Risk(s_{k,l},\theta; \varphi_{k.l})
$$
Therefor,
$$
Risk(s_{i,j}, \theta; \varphi_{i,j}) \leq Risk(s'_{i,j}, \theta; \varphi_{i,j})
$$
It implies that,
$$
a_{i,j}P_{\theta}(\varphi_{i,j}=1) \leq b_{i,j}P_{\theta}(\varphi_{i,j}=0), \ \mbox{if} \ \theta \in \omega_{i,j} 
$$
and:
$$
b_{i,j}P_{\theta}(\varphi{i,j}=0) \leq a_{i,j}P_{\theta}(\varphi_{i,j}=1), \ \mbox{if} \ \theta \in \omega^{-1}_{i,j} 
$$
This means that the individual test $\varphi_{i,j}$ satisfies (\ref{unbiased_individual})  with the significance level $\alpha_{i,j}=b_{i,j}/(a_{i,j}+b_{i,j})$. 
Taking into account that $\varphi^{umpu}_{i,j}$ is optimal in the class of unbiased tests one gets
$$
Risk(s_{i,j}, \theta; \varphi^{umpu}_{i,j}) \leq Risk(s_{i,j}, \theta; \varphi_{i,j})
$$ 
Therefore,
$$
Risk(S, \theta;\delta^{ou})= \sum_{i=1}^p\sum_{j=1}^p Risk(s_{i,j},\theta; \varphi^{umpu}_{i,j}) \leq 
$$
$$\leq \sum_{i=1}^p\sum_{j=1}^p Risk(s_{i,j},\theta; \varphi_{i,j})=Risk(S, \theta;\delta),
$$
for any $w$-unbiased statistical procedure $\delta$. The theorem is proved.

The main result of the paper is the following
\begin{teo}
Let $0 < \alpha <1$, and the loss function $w$ be defined by (\ref{additive_loss_function}) with $a_{i,j}=1-\alpha$, $b_{i,j}=\alpha$, $i,j =1,2,\ldots,p$. Then for any $w$-unbiased multiple decision statistical procedure $\delta$ defined by (\ref{test_for_N_hypotheses_overall_form})- (\ref{mdp_overall_form}) one has
$$
(1-\alpha)E_{\theta}[Y_I(S, \delta^{ou})]+\alpha E_{\theta}[Y_{II}(S, \delta^{ou})] \leq (1-\alpha)E_{\theta}[Y_I(S, \delta)]+\alpha E_{\theta}[Y_{II}(S, \delta)], 
$$
for any $S \in \cal{G}$, $\theta \in {\Omega}_S$. Here $Y_I(S, \delta)$, $Y_{II}(S, \delta)$ are the numbers of Type I and Type II errors for model selection by statistical procedure $\delta$.
\end{teo}

\noindent
{\bf Proof.} For any statistical procedure $\delta$ one has from the Theorem 1
$$
Risk(S, \theta;\delta)=(1-\alpha)E_{\theta}[Y_I(S, \delta)]+\alpha E_{\theta}[Y_{II}(S, \delta)]
$$
From Theorem 2 one has:
$$
Risk(S, \theta;\delta^{ou}) \leq Risk(S, \theta;\delta)
$$
for any $w$-unbiased procedure $\delta$. The theorem is implied. Note that in this case $\alpha$ is the significance level of all individual tests.  

\section{Concluding remarks}\label{Concluding remarks}
The main result of the paper states that statistical procedure $\delta^{ou}$ gives the minimal value for the linear combination of expectations of numbers of Type I and Type II errors 
for any true Gaussian graphical model $S$. It is interesting to compare  Gaussian graphical model selection procedures  known in literature with respect to this criteria for different 
$S$. Our experiences shows that the structure of $S$ will play an important role in this comparison. This will be a subject for a forthcoming study.  

\noindent
{\bf Aknowledgement:} This work is conducted at the National Research University Higher School of Economics,  Laboratory of algorithms and technologies for network analysis. The authors Kalyagin V. and Pardalos P. are supported by RSF grant 14-41-00039.

\end{document}